# Improving Streaming Video Segmentation with Early and Mid-Level Visual Processing

Subarna Tripathi[1], Youngbae Hwang[2], Serge Belongie[1,3], Truong Nguyen[1]

[1]University of California San Diego, [2]Korea Electronics Technology Institute, [3]Cornell NYC Tech

## Abstract

*Despite recent advances in video segmentation, many opportunities remain to improve it using a variety of low and mid-level visual cues. We propose improvements to the leading streaming graph-based hierarchical video segmentation (streamGBH) method based on early and mid level visual processing. The extensive experimental analysis of our approach validates the improvement of hierarchical supervoxel representation by incorporating motion and color with effective filtering. We also pose and illuminate some open questions towards intermediate level video analysis as further extension to streamGBH. We exploit the supervoxels as an initialization towards estimation of dominant affine motion regions, followed by merging of such motion regions in order to hierarchically segment a video in a novel motion-segmentation framework which aims at subsequent applications such as foreground recognition.*

## 1. Introduction

Video analysis literature has two predominant research directions in feature-based [1] methods and segmentation-based methods [2]. Several driving applications such as object recognition [3], augmented reality [4] or animation [5] require dense segmentation, and feature based method is not a natural fit in such cases. Thus video segmentation acquired a rich history in the last decade.

Despite of several approaches proposed in the literature of video segmentation, more recently the idea of associating initial over-segmentations into supervoxels, followed by processes such as hierarchical grouping [6], long range tracking [7], superpixel flow [8] have been proposed. As an early vision tool, color-based supervoxel segmentation method "streaming graph-based hierarchical segmentation" (streamGBH) [9] reportedly has the best performance in terms of quality and complexity. In this paper we propose a significant increment and improvement to the state-of-the-art streamGBH in terms of improving segmentation quality by using dense optical flow. We use optical flow as feature and at the same time as a guiding tool for the temporal connection in the initial graph. We perform thorough experimental analysis on a benchmark database [10] used in the evaluation of libsvx library [11] for streamGBH. We evaluate our approach visually and in terms of video segmentation objective metrics. We also discuss further extension of streamGBH towards the challenging task of video analysis. We exploit the supervoxels as an initialization for the estimation of dominant affine motion regions followed by merging of such motion regions based on their geometric distance. We present a novel framework of hierarchical motion based video segmentation to enable a powerful intermediate level video representation for subsequent recognition or other task-specific applications.

The remainder of this paper is as follows. In Section 2, we outline related work. The proposed video segmentation methodology using early and mid-level visual processing is presented in section 3. Experimental results are reported and commented in section 4, followed by concluding remarks in section 5.

## 2. Related Work

There are three different paradigms in video segmentation. First is frame processing in which each frame is independently segmented, but no temporal information is used. This method is fast but the temporal coherence is poor. Second is 3D volume processing that represents a model for the whole video. It is bi-directional multi-pass processing. The results are best, but the complexity is too high to process long videos and does not cater to the need for streaming videos. Stream processing processes the current frame only based on a few previously processed frames. It is forward-only online processing, and the results are good and efficient in terms of time and space complexity. The state-of-the-art streaming segmentation (streamGBH) outperforms other streaming methods and competitive with full-video hierarchical methods. The streamGBH in libsvx implements this video segmentation approximation framework. In this framework, the streaming video is conceptualized as a set of non-overlapping subsequences $\mathrm{v} = \{v_1, v_2, \ldots, v_m\}$ with $k_i$ frames for subsequence $v_i$. The hierarchical segmentation result, $s$, is approximately decomposed into $\{s_1, s_2,\ldots,s_m\}$, where $s_i$ is hierarchical segmentation of subsequence $v_i$. StreamGBH adopts the graph-based grouping method into the streaming hierarchical segmentation framework. It constructs a graph which is similar to a graph over the spatial-temporal video volume with a 26-neighborhood in 3D space-time. However, this graph is only constructed for the current two subsequences in process, $v_i$ and $v_{i-1}$. This graph is the first layer of the hierarchy and its edge weights are direct color dissimilarity (measured by $\chi^2$ distance of the normalized color histogram) of voxels. The streaming hierarchical segmentation results upto subsequence $v_i$, uses ideas from the previous section. In other words, given the hierarchical segmentation result $s_{i-1}$ of $v_{i-1}$, the layer by layer hierarchical segmentations for $v_i$ are inferred. Being a streaming segmentation method, StreamGBH thus becomes a powerful early vision tool.

When the question of target applications such as animation or object recognition or 2Dto3D video conversion arises, we need to go beyond the early vision problem of supervoxel segmentation. In order to do task-specific jobs such as recognizing objects, we intend to achieve hierarchical motion layer representations of a video such that a single segment would represent a single object. Layered models offer an elegant approach to motion segmentation and have many advantages. A typical scene consists of very few moving objects and representing each moving object by a layer allows the motion of each layer to be described more simply [12]. Though motion analysis on longer image sequences [13] can produce best results; however, the computational and memory complexity becomes an issue for online applications. There are several methods for pursuing motion-based layer segmentations such as [12]. However, we are not aware of any method that does streaming hierarchical motion-layer segmentation.

# 3. Methodology

The proposed method for video segmentation uses early and mid-level visual processing. We improve streamGBH by incorporating edge-preserving smoothing and motion information. We also extend streamGBH for hierarchical motion segmentation.

## 3.1. StreamGBH+

As a pre-processing step, we apply bilateral filtering [14] as an edge-preserving smoothing to improve segmentation. Along with color similarity, we consider motion similarity of voxels and influence of motion direction on graph connectivity. Use of dense optical flow [15] considerably improves segmentation results. Firstly, instead of connecting a voxel $(i,j,t)$ to its immediate 9 neighbors $(i+m, j+n,t-1)$, $m,n \in \{-1,0,+1\}$ in the previous frame, we connect it to its 9 neighbors along the backward flow vector $(u,v)$, i.e. $(i+u(i,j)+m, j+v(i,j)+n, t-1)$ similar to the approach proposed in [6]. This is a generalization of prior grid-based volumetric approaches which can only be achieved using a graph representation.

Secondly, we use optical flow as a feature for each region during hierarchical segmentation. As optical flow is only consistent within a frame, we use a per-frame discretized flow histogram. Unlike [6] which discusses SIFT-like (with respect to angle) motion feature representation, we propose simpler representation of two histograms of horizontal and vertical component of optical flow field. The benefit of this simpler approach is to distinguish motions with same direction but different magnitude. Matching the flow-descriptors of two regions then involves averaging the $\chi^2$ distance of their normalized per-frame flow-histograms over time.

We combine the $\chi^2$ distance of the normalized color histograms $d_c \in [0,1]$ with the $\chi^2$ distance of the normalized flow histograms $d_f \in [0,1]$ by

$$(d_c, d_f) \rightarrow (1-(1-d_c)(1-d_f))^2 \qquad (1)$$

This function is close to zero if both distances are close to zero, and close to one if any one of them is close to one. We also tried exponential form of distance function which produces similar result.

Throughout this paper we term our approach of supervoxel segmentation as streamGBH+.

## 3.2. Supervoxel to hierarchical motion-layer segmentation

We build a hierarchical motion layer framework on top of streamGBH+. We estimate the motion of each segment and hierarchically merge different segments using geometric distance notion in affine space in a streaming fashion.

### *3.2.1 Motion-based segmentation overview*

Following the approach of most of the layered segmentation techniques, we also assume a parametric motion for each layer. Simple translation or just a rotation might be too restrictive to capture the motion of natural scenes; so we consider affine motion. We process optical flow for frame-pairs and estimate affine parameters in streaming fashion and propose a hierarchical online motion segmentation framework. We see different aspects of the motion scene in the layers of hierarchy. Different regions following affine motions are merged to produce smaller number of foreground objects as the hierarchy level increases. The novel framework for merging spatial regions is based on geometric distance conceptualized as directed divergence in affine space.

### *3.2.2 Segmentations and streamGBH+*

Affine motion parameter estimation in regions iterate over hierarchy levels. Initialization at the lowest level of motion hierarchy is influenced by the output from streamGBH+. Unlike the block-based estimation of affine parameters [12], we initialize affine parameter estimation from optical flow field using RANSAC over the segments from streamGBH+. Thus every region has an initial parametric model estimated. The most important question then becomes which regions are to be merged at what level of hierarchy. To address this, we take the notion of geometric distance between affine regions. The dissimilarities between affine regions directly depend on the directed divergence described below from associated warping error between those regions. As shown in Figure 1, at a certain level of motion hierarchy, every region has an affine motion parameter set associated with it. Clearly the torso of the player, his legs, the tennis racket, two parts of the gallery are in different segments.

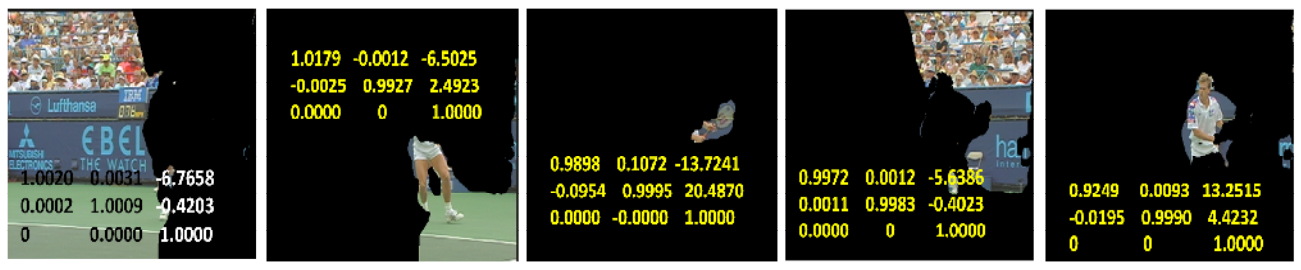

Figure 1: Different regions demonstrated in non-black colors with their corresponding affine motion parameters

We aim at estimating motion layers by merging supervoxels or over-segmented regions having apparently little different spatio-temporal feature. We propose a criterion to use for merging two regions based on affine flow. The geometric distance between affine regions uses the notion of warping error based dissimilarity, directed divergence. Directed divergence of region $\#k$ from region $\#i$ $div(k,i)$ and directed divergence of region $\#i$ from region $\#k$ $div(i,k)$ are defined as:

$$\left.\begin{aligned} div(i,k) &= \frac{i}{n_i}\sum\left[M1_{rc} = abs\left(TC\left(T_i(R_i)\right) - TC\left(T_k(R_i)\right)\right)\right] \\ div(k,i) &= \frac{i}{n_k}\sum\left[M2_{rc} = abs\left(TC\left(T_k(R_k)\right) - TC\left(T_i(R_k)\right)\right)\right] \end{aligned}\right\} \quad (2)$$

Where, $\Sigma$ denotes matrix-wide summation i.e. sum of all elements of a matrix; $R_i$ and $R_k$ are $i$-th and $k$-th region. $T_i$ is the transformation estimated for region #i. In order to have a fair comparison base for different regions of different sizes, we transform every region with respect to a canonical reference frame i.e. the transformation to a pre-determined image-size (say, p x q). TC is the transformation to canonical reference frame which always yields a matrix of size p x q. The sum over all the elements of the difference matrix is then normalized by the corresponding number of pixels in a region e.g. $n_i$ is the number of pixels in region #i and $n_k$ is the number of pixels in region #k. Figure 2 shows distance evaluation visually. The maximum among the pair *(i,k)* and *(k,i)* of directed divergence values is treated as the distance between two regions in their affine space. The distance metric is defined as follows:

$$dist(i,k) = dist(k,i) = \max\left(div(i,k), div(k,i)\right) \quad (3)$$

Two regions can be merged if the distance between the corresponding affine spaces is small. If either of *div(i,k)* or *div(k,i)* is high, the region *#i* and region *#k* are not merged.

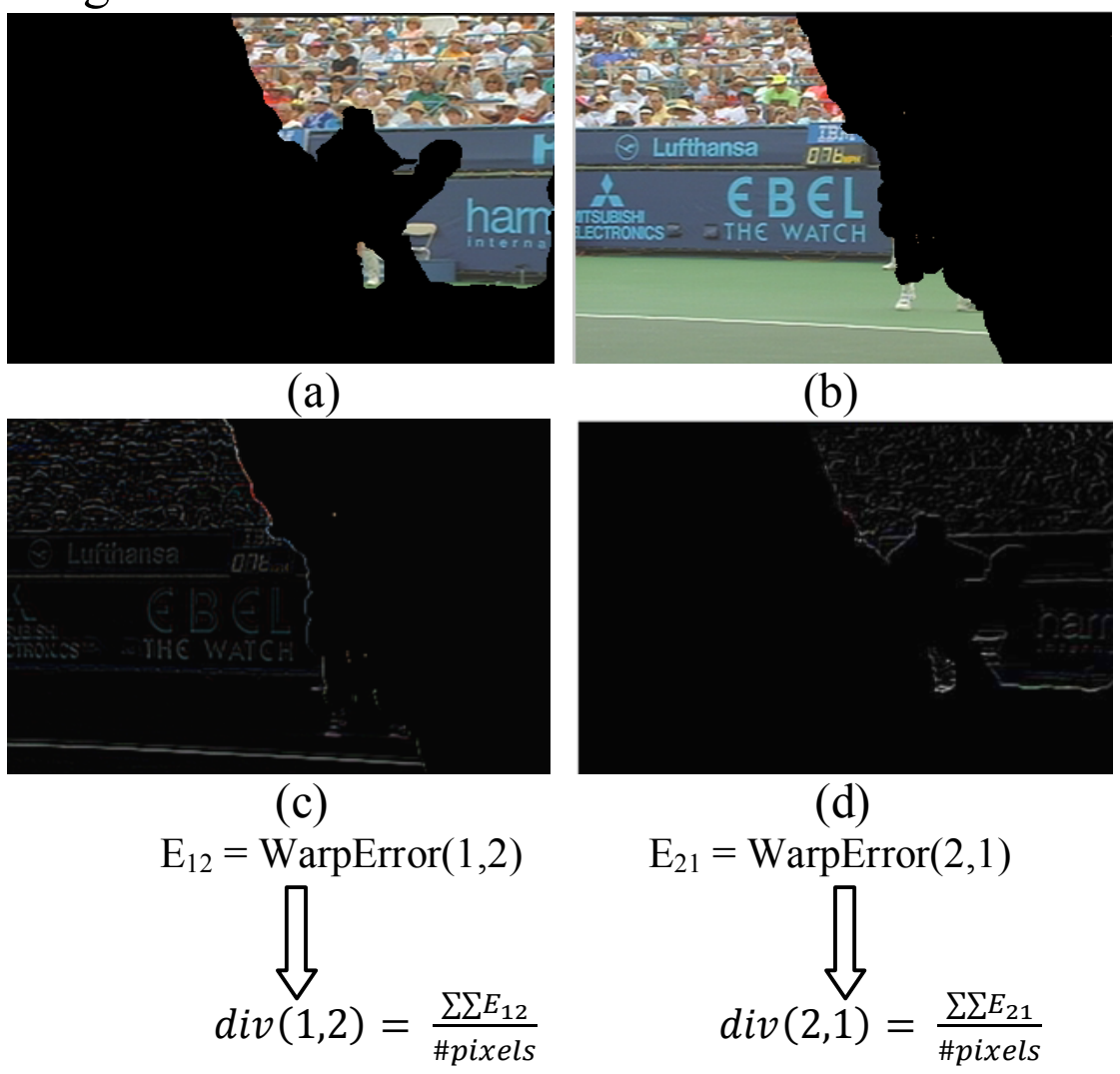

Figure 2: Evaluating geometric distance between (a) affine regions#1 and (b) affine region #2. Apply following transformations to region #1. Affine transformation associated with region #1, followed by a transformation into canonical reference frame (fixed size). Similar operations for region number #1with transformation #2. (c) Normalized sum of pixel-level warping error, $div(1,2)$, is the directed divergence of region 2 from region 1. (d) Similarly $div(2,1)$ is evaluated. Clearly, $div(1,2)$ is not same as $div(2,1)$

### *3.2.3 Grouping method at layer hierarchies*

We consider first fit or local greedy way as the region grouping method. Our assumption is if region i, region j, and region k all are to be merged; the merging criteria for all of the pairs *(i,j)*, *(j,k)* or *(i,k)* would be satisfied. No matter with which pair we start, eventually we would come to the same grouping at the end. The directed divergence and distance metric concept has been visualized in Figure 3. Directed divergence metric is not symmetric. As explained in the figure, $div(1,2)$ and $div(2,1)$are not same.

After the region merging, the significant errors are found especially near the object boundary. To overcome this problem, we explore Markov Random Field based smoothing as described in [16]. As we know the number and parameters for candidate motions at each hierarchy level, we apply BVZ [17] algorithm to achieve smoother segments. In the MRF framework, we intend to optimize both the pixel-wise warping error satisfying a prior smoothness constraint.

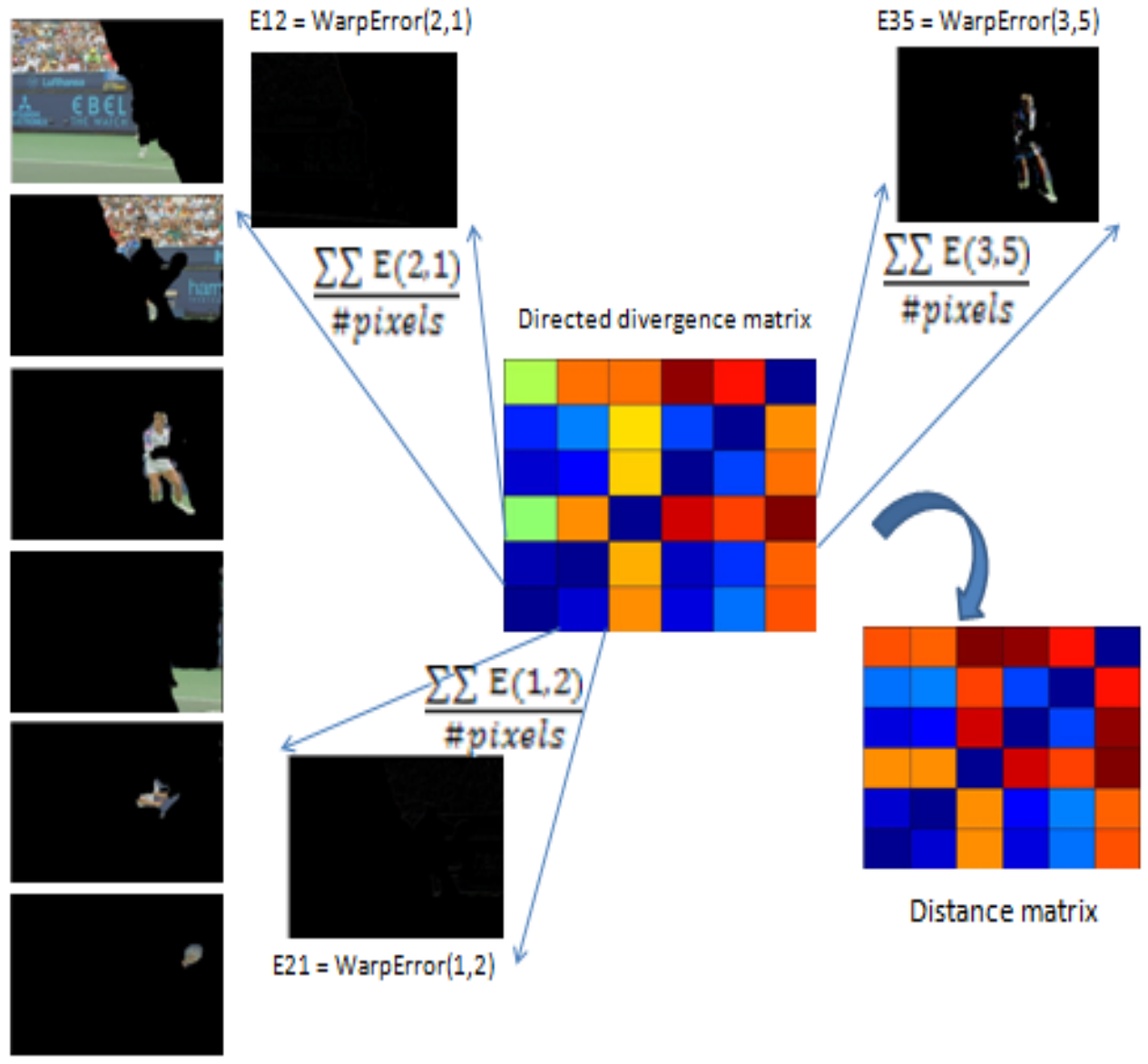

Figure 3: Geometric distance based affine region merging: six initial regions are shown in left most border. The square directed divergence matrix is shown in the center. Each element in this matrix is based on normalized sum of directed warping errors. An element of the symmetric distance matrix is the maximum between a pair of directed divergence entries. Entries (1,2), (2,1) and (3,5) in directed divergence matrix have been visualized as examples. Color code: higher color temperature means higher distance values. The diagonal entries in the matrix mean distance with self, which are zeros. The warm color temperature e.g. entry for $div(3{,}5)$ means high distance value

### *3.2.4 Temporal consistency*

Enforcing temporal consistency has two paradigms. Firstly, since the affine motion is consistent within a frame, we cannot use their values as we do in volume graph using color feature value. Rather, we use a per-frame discretized flow histogram and thus the association of a motion segment from previous to current frame is not inherent. To address the issue of enforcing temporal consistence in simple and effective way, we use the segmentation from backward warping from the previous frame-pair as an initialization for the segmentation process (forward warping) of current frame-pair. Thus, the question of associativity between segments over time becomes only about an associativity of segments from forward warping to backward warping of a frame-pair.

The best matching segment, in terms of overlapping area, in the second image (wrt the warped segment in the first image) establishes segment correspondence between forward and backward warped motion segmentations.

## 4. Experimental Evaluations

We demonstrate the results of applying streamGBH+ and hierarchical motion segmentation onto several videos.

## 4.1. streamGBH+

### *4.1.1 Quantitative Performance: Benchmark Comparisons*

**Data:** We use the recently published benchmark dataset (ChenXiph.org) [10] and video segmentation performance metrics [11] for the quantitative experiments. This video dataset is a subset of the well-known xiph.org videos that have been supplemented with a 24-class-semantic-pixel labeling set (The same classes from the MSRC object-segmentation dataset [10]). In the implementation, we use sequence length 3 in all experiments and thus performance is not expected to be near to full-video processing.

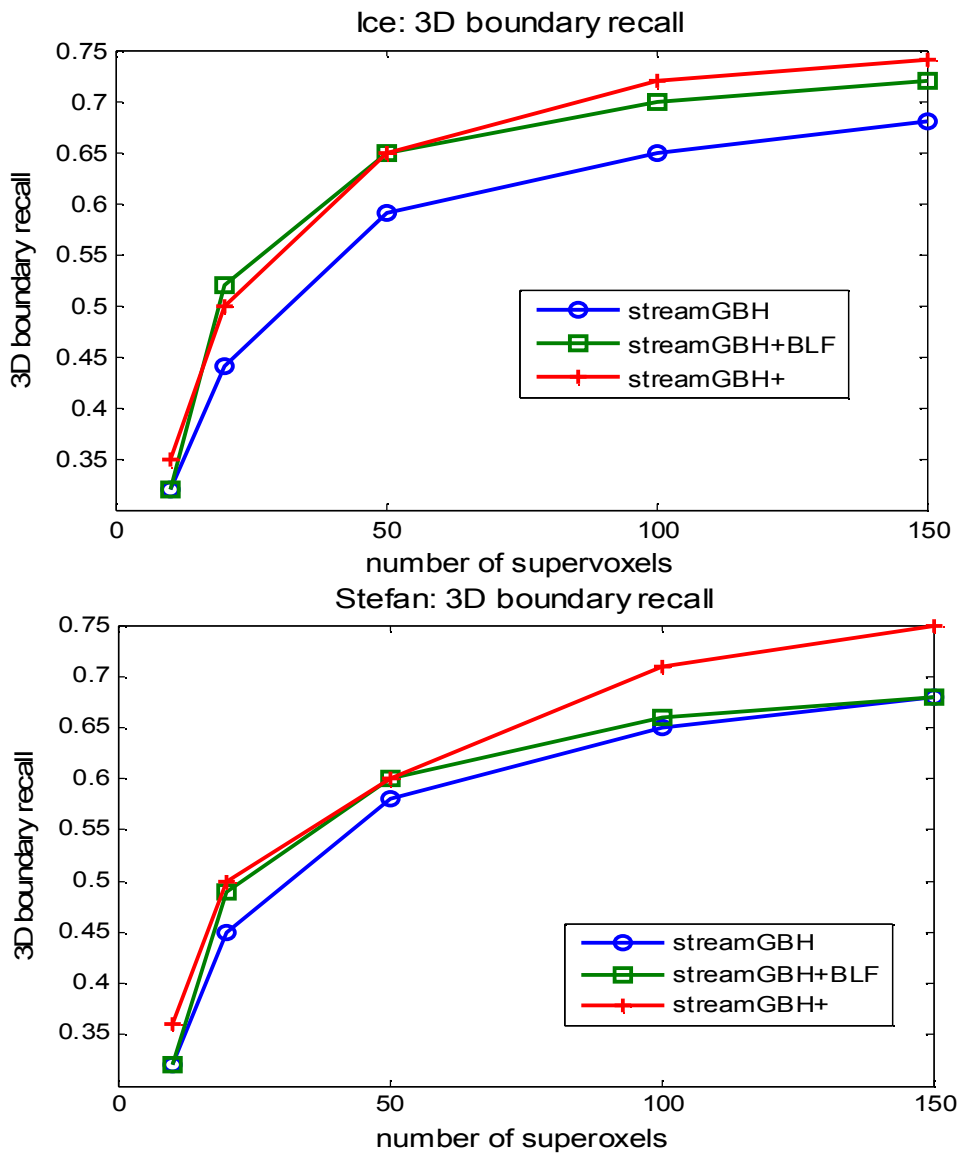

Figure 4: number of supervoxel vs. 3D Boundary Recall metric values for streamGBH, streamGBH with bilateral filtering and streamGBH+ (streamGBH+ bilateral filtering+ motion). Top: evaluation result for the sequence Ice and bottom: evaluation result for the sequence Stefan.

The 8 videos ('Bus', 'Container', 'Garden', 'Ice', 'Paris', 'Salesman', 'Soccer' and 'Stefan') in this set are densely labeled with semantic pixels and have duration of 85 frames each. This dataset has been used for evaluation for the state-of-the-art StreamGBH method. This dataset allowed us to evaluate these segmentation methods against human perception.

**3D Boundary Recall:** The 3D boundary is the shape boundary of a 3D object, composed by surfaces. It measures the detection of spatio-temporal boundary. For each segment in the ground-truth and segmentations, we extract the within-frame and between-frame boundaries and measure recall using the standard formula [11]. Figure 4 shows the dependency of 3D Boundary Recall on the number of segments. StreamGBH+ performs better.

**Explained Variation:** Explained Variation metric is proposed in [18] as a human-independent metric. It considers the supervoxel as a compression method of a video. The metric is defined as:

$$R^2 = \frac{\sum_i(\mu_i-\mu)^2}{\sum_i(x_i-\mu)^2} \quad (3)$$

It is evaluated by summing over $i$ voxels where $x_i$ is the actual voxel value, μ is the global voxel mean and $\mu_i$ is the mean value of the voxels assigned to the supervoxel that contains $x_i$ [11].

Figure 5 shows the dependency of explained variation metric on the number of supervoxels. StreamGBH+ again performs better than streamGBH.

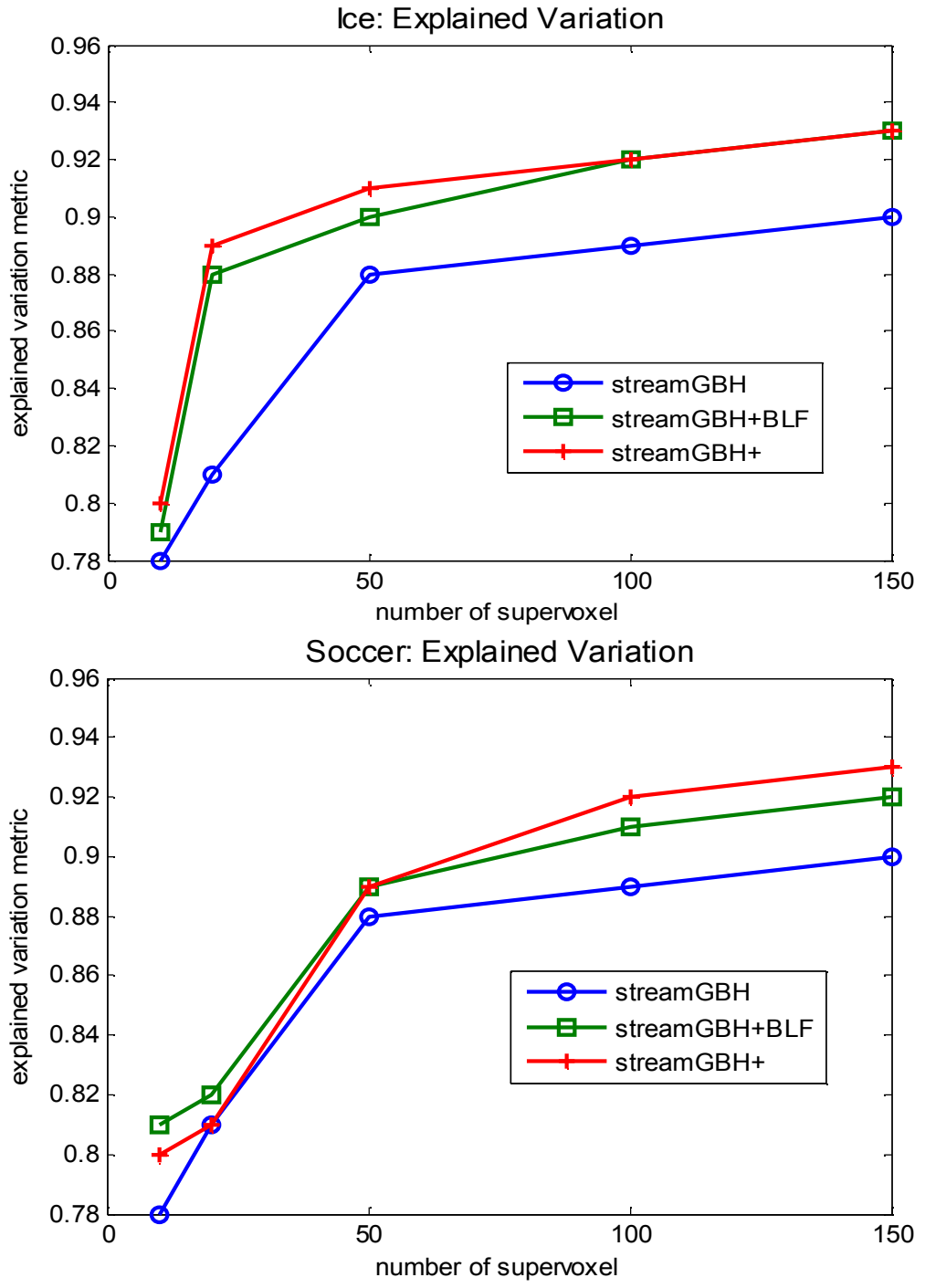

Figure 5: Number of supervoxel vs. Explained Variation metric values for streamGBH, streamGBH with bi-lateral filtering and streamGBH+. Top: evaluation result for the sequence ICE and bottom: evaluation result for the sequence Soccer

**3D Segmentation accuracy:** This metric measures what fraction of a ground-truth segment is correctly classified by the supervoxels; each supervoxel should overlap with only one object/segment as a desired property [11] of video segmentation. To evaluate the overall segmentation quality, we also take the average of the fraction across all ground-truth segments in the video. Figure **6** shows the dependency of 3D accuracy on the number of supervoxels. Here also streamGBH+ performs better than streamGBH.

**3D Under-segmentation Error:** 3D under-segmentation error measures what fraction of voxels goes beyond the volume boundary of the ground-truth segment when mapping the segmentation onto it. The details for this metric can be found in the benchmark paper [11].

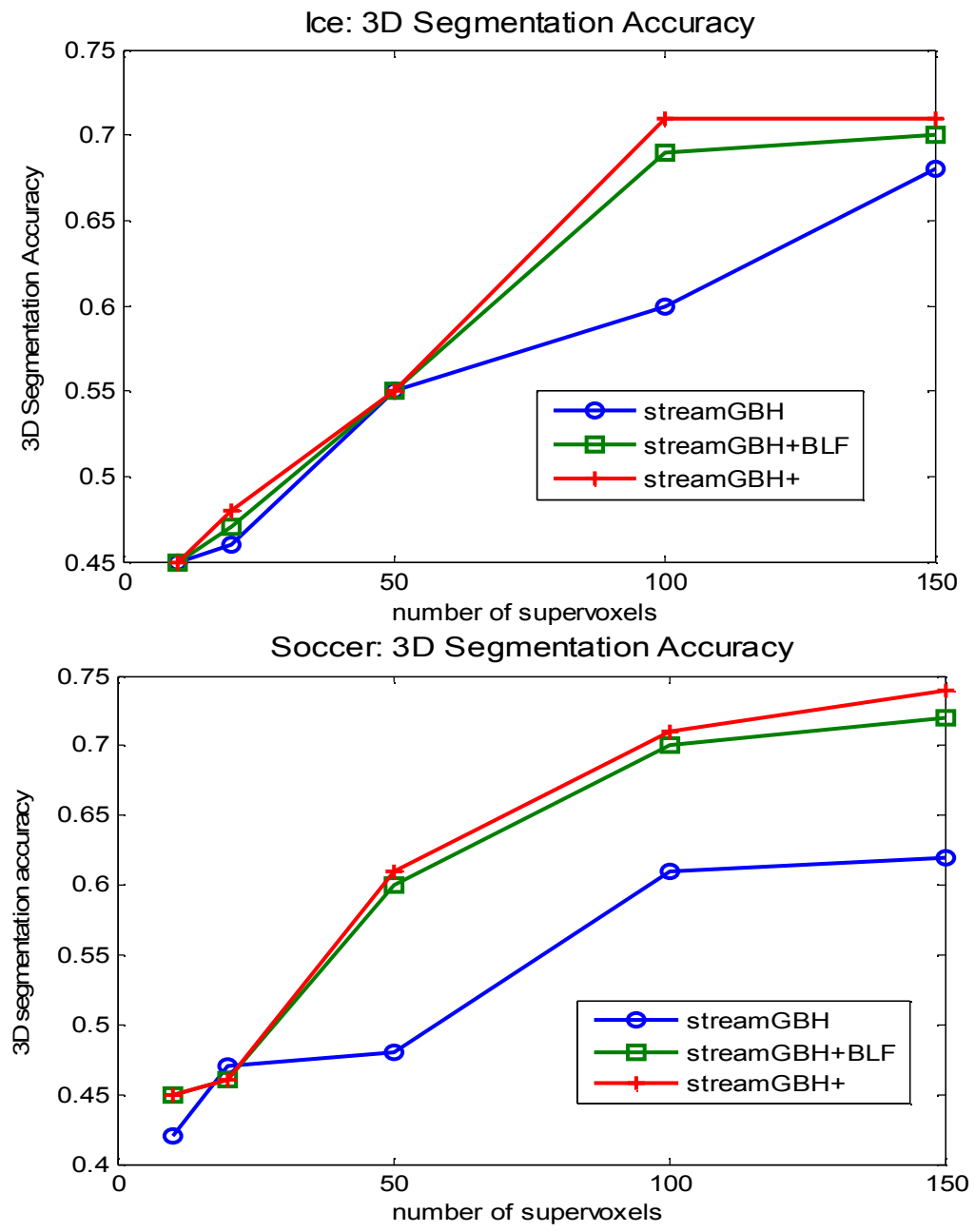

Figure 6: number of supervoxel vs. 3D segmentation accuracy metric values for streamGBH, streamGBH with bi-lateral filtering and streamGBH+ (streamGBH + bi-lateral filtering + motion). Top: evaluation result for the sequence ICE and bottom: evaluation result for the sequence Soccer

| **metric** | **streamGBH** | **streamGBH + motion** | **streamGBH + BLF** | **streamBHG+** |
|---|---|---|---|---|
| boundary recall 2D | 0.44 | 0.442 | 0.451 | **0.452** |
| boundary recall 3D | 0.47 | 0.482 | 0.49 | **0.49** |
| explained variation | 0.71 | 0.72 | 0.73 | **0.75** |
| accuracy 2D | 0.58 | 0.58 | 0.58 | 0.58 |
| accuracy 3D | 0.56 | .54 | 0.55 | 0.55 |
| Under segmentation error 2D | 8 | 9 | 9 | 10 |
| Under-segmentation error 3D | 18 | 18 | 15 | 18 |

Table 1: Average comparative metric values for all the videos in Chen database

Proposed streamGBH+ performs better than the state-of-the-art streamGBH for videos with object motions. Overall, streamGBH+ outperforms in the evaluation of metrics such as boundary recall 2D, boundary recall 3D, explained variation. As per Table 1 for the metrics accuracy 2D and accuracy 3D we see no difference for this database overall. Though, the performance of streamGBH+ for the under segmentation metrics is not better than the state-of-the-art, we care for other metrics more at this time because of our future target application on object recognition or 2D-to-3D video conversion.

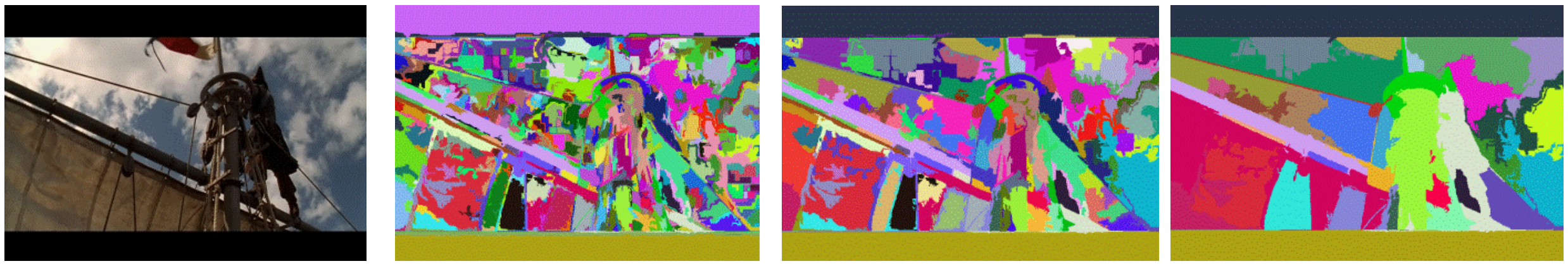

Figure 7: Different details of objects in the layers of hierarchy. Video frame, supervoxel segmented frame at hierarchy level 1, 10 and 15

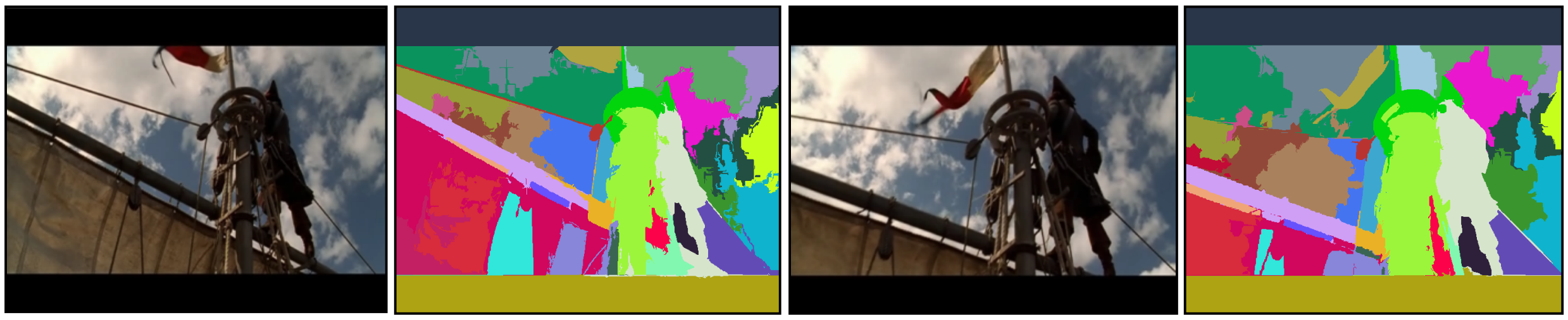

Figure 8: Qualitative time consistent performance of streamGBH+. Left to right: pairs of video frame and supervoxel segmented frame from video clip "Pirate".

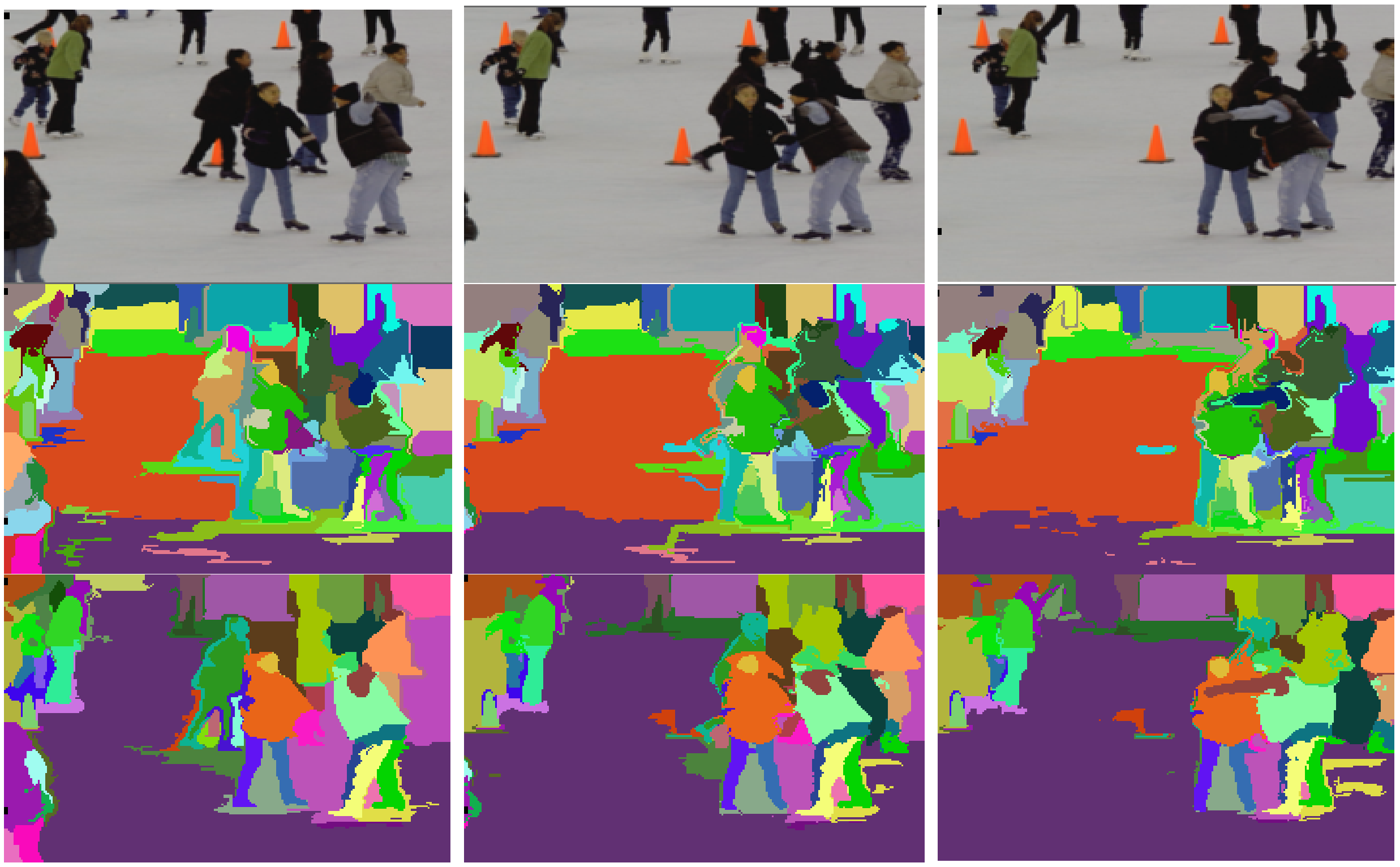

Figure 9: Qualitative performance of streamGBH+. Top: input video "Ice", frame 1, frame 15 and frame 26. Middle: segmentation result of state-of-the-art streamGBH at hierarchy level 5. Bottom: segmentation result of our method streamGBH+ at hierarchy level 5

We use the optical flow method **[15]** based on constant memory; coarse to fine warping techniques which uses fixed point iterations. Thus the overhead of memory and computational complexity of streamGBH+ is not significant compared to streamGBH.

**Qualitative Performance:** Here we show some qualitative results on long videos, which necessitate a streaming method. We see different details of an object in the layers of hierarchy. For example, in Figure 7 one can see more than three parts in the object "Pirate" in 5th layer, and locate a single human body pose in 15th layer. Figure 8 shows a long term temporal coherence of streamGBH+. Figure 9 corroborates that streamGBH+ is able to avoid unnecessary over-segmentation compared to state-of-the-art streamGBH.

## 4.2. Motion Segmentation Results

As we increase the hierarchy level, we allow higher acceptance for distances between affine regions for merging feasibility and thus more number of regions are combined. Figure 10 and Figure 11 demonstrate the results on sequence Stefan and Lovebird. As some regions are merged, we refine our affine parameters by estimating them again over the merged regions.

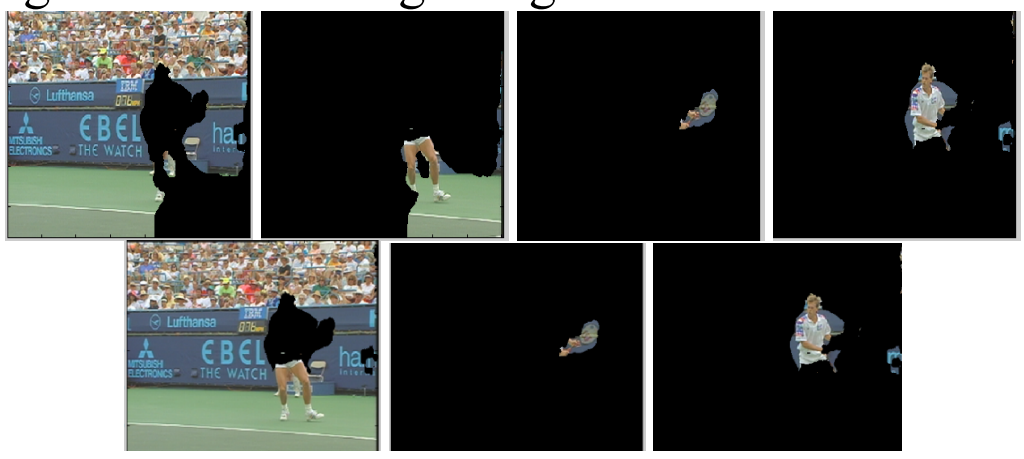

Figure 10: Two layers of hierarchical motion segmentation on Stefan. Top: Hierarchy level 1 has 4 regions; the background, the legs of the player; the tennis racket and the torso. Bottom: Level 2 has three layers

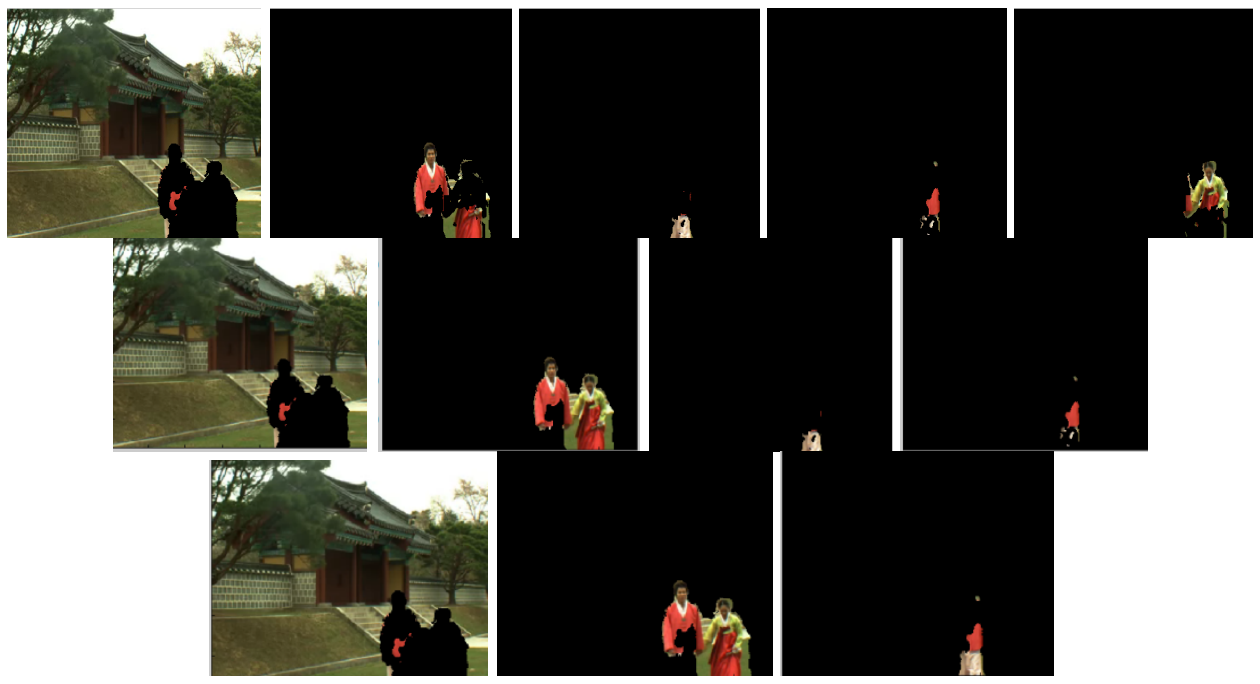

Figure 11: Three layers of hierarchical motion segments on sequence Lovebird. Top: Hierarchy level 1 has 5 regions; middle: h-level 2 has 4 regions; bottom: h-level 3 has 3 regions

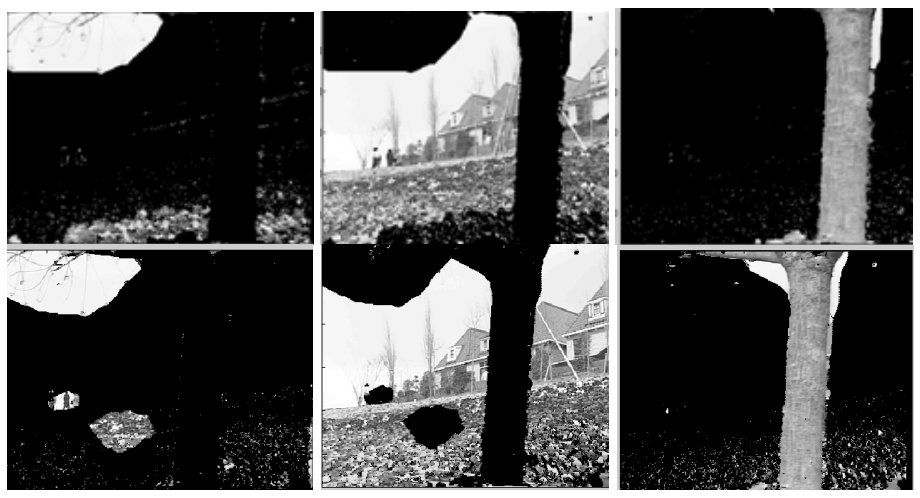

Figure 12: Temporal consistency shown for hierarchy level 2 motion layer segmentation of sequence Garden. Top: 3 segments estimated for frame #1. Bottom: corresponding 3 layers for frame #10

Figure 12 shows one example of the temporal consistency of our streaming motion segmentation. Here, we show segmentation at hierarchy level 2 for frame number 1 and frame number 10. Note the temporal correspondence between layers. Some more results for hierarchy level 3, where each segment is represented in separate color, have been shown Figure 13. In all these videos except the "Calendar" sequence, the camera is moving, thus the background is not static.

In the first case, the car and the background are segregated based on motion analysis. In the second case, the background, the car and the cyclist have been discriminated. In sequence "Calendar", the ball and the toy engine are distinguished from the static background. In the "Garden" sequence the tree segment has been segregated but a little part of the flower bed has been merged with the tree at this hierarchy level. In the "Lovebird" sequence, two persons are assigned different motion layers compared to the background. In Stefan, the tennis player, his racket and the background (including the gallery and the court) are detected as different layers in this hierarchy. For most of the cases, object level motion layer has been detected at one hierarchy level or other.

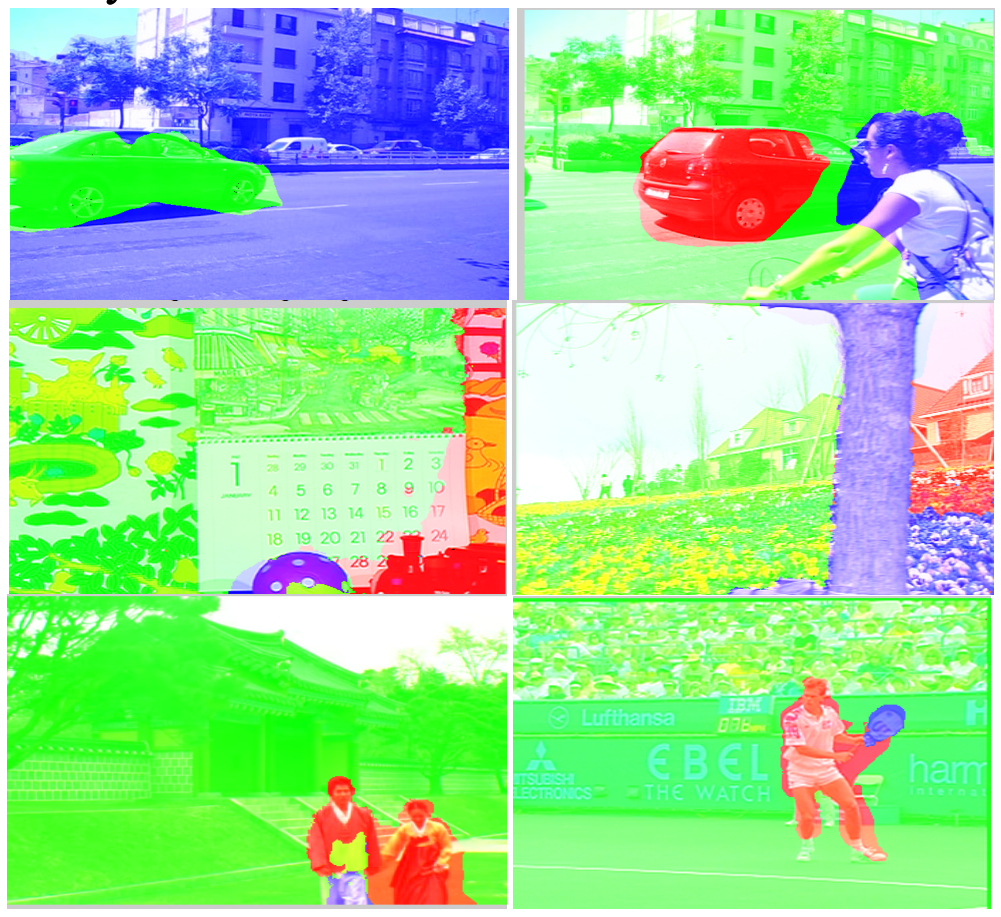

Figure 13: hierarchy level 3 motion layer segmentation. Top to bottom and left to right: results on Car1, Car2, Mobile, Garden, Lovebird, Stefan.

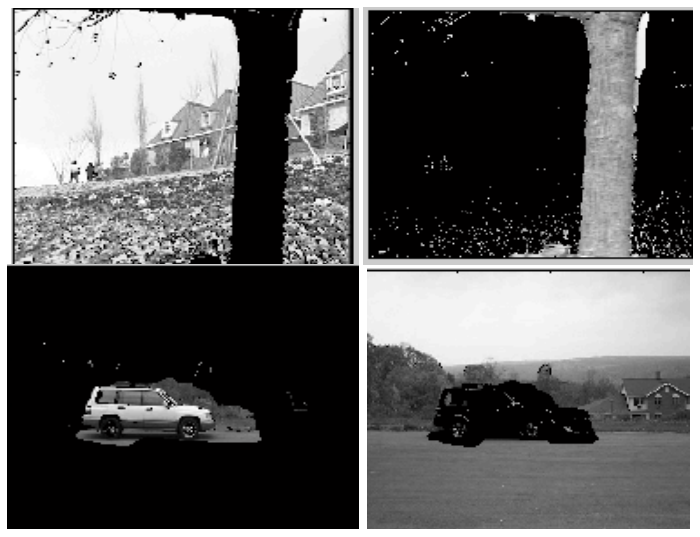

Figure 14: segmentation after MRF smoothing at hierarchy level 3. Top: Motion layer segmentation of Garden, Bottom: motion layer segmentation of Car

In order to smooth out the segmentations and reduce the error near object boundaries, Markov Random Field based post-processing has been applied. Figure 14 shows MRF based smoothing results on sequence Garden and Car. However, as we apply MRF based smoothing on Stefan, we tend to miss the small region of different motion of the tennis racket. Though the segmentation around the torso of the player becomes stricter, but the motions of the lower part of the body of the player and the tennis racket have been merged with the background.
Thus, MRF isn't always a win as shown in Figure 15.

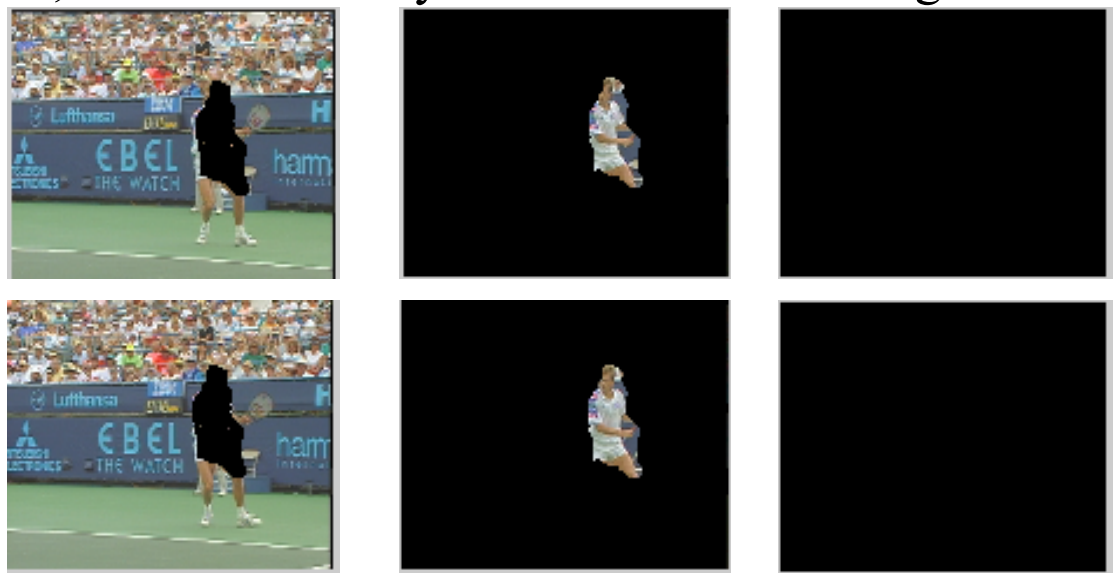

Figure 15: segmentation after MRF smoothing. Top: segmentation from forward warping. Bottom: segmentation from backward warping. Segment of tennis racket (before MRF as in Figure 13) does not have any associated regions for both forward and backward warping after MRF

## 5. Concluding remarks

We propose improvement towards hierarchical video segmentation using early and mid-level visual processing.

We have presented an improvement to the state-of-the-art color-based supervoxel segmentation methodology, called "Streaming Graph-based Hierarchical" segmented method by incorporating motion information. We have evaluated the performance of our approach vis-à-vis the performance of streamGBH on different videos and have shown that our approach, streamGBH+, produces improved supervoxels compared to the state-of-the-art.

We also propose an extension to streamGBH+ from supervoxel to layered representation. We exploit the supervoxels as an initialization for the estimation of dominant affine motion regions followed by merging of such motion regions based on their geometric distance.

We plan to evaluate the performance of the streaming hierarchical motion-layer segmentation on other motion analysis database in near future.

**Acknowledgements:** This work is supported by the Technology Development Program for Commercializing System Semiconductor funded by the Ministry Of Trade, Industry and Energy (MOTIE, Korea). (No 10041126, Title: International Collaborative R&BD Project for System Semiconductor)